# The Application of Deep Convolutional Networks to Ultrasound for Modeling of Dynamic States within Human Skeletal Muscle

Ryan J. Cunningham, Member, *IEEE*, Peter J. Harding, and Ian D. Loram, *Member, IEEE*

*Abstract*— This paper concerns the fully automatic direct *in vivo* measurement of active and passive dynamic skeletal muscle states using ultrasound imaging. Despite the long standing medical need (myopathies, neuropathies, pain, injury, ageing), currently technology (electromyography, dynamometry, shear wave imaging) provides no general, non-invasive method for online estimation of skeletal intramuscular states. Ultrasound provides a technology in which static and dynamic muscle states can be observed non-invasively, yet current computational image understanding approaches are inadequate. We propose a new approach in which deep learning methods are used for understanding the content of ultrasound images of muscle in terms of its measured state. Ultrasound data synchronized with electromyography of the calf muscles, with measures of joint torque/angle were recorded from 19 healthy participants (6 female, ages: 30 ± 7.7). A segmentation algorithm previously developed by our group was applied to extract a region of interest of the medial gastrocnemius. Then a deep convolutional neural network was trained to predict the measured states (joint angle/torque, electromyography) directly from the segmented images. Results revealed for the first time that active and passive muscle states can be measured directly from standard b-mode ultrasound images, accurately predicting for a held out test participant changes in the joint angle, electromyography, and torque with as little error as 0.022°, 0.0001V, 0.256Nm (root mean square error) respectively.

## I. Introduction

There is a current unmet medical demand for personalized *in vivo* skeletal muscle analysis. Neurological conditions (dystonia, motor neurone disease), myopathies (myositis, in-flammation), neuropathies (nerve injury, spinal cord injury), ageing (motor unit loss), and pain/injury (work-related injury, neck injury, back injury, low back pain, neck pain) are some medical problems which would benefit from an ability to measure the dynamic/static states of specific/individual skeletal muscles *in vivo*. The state of muscle is determined by numerous input factors, the main two being neural drive, and joint rotation. Active contraction/relaxation via neural drive causes muscle shortening/lengthening. Muscles can lengthen or shorten when connecting joints rotate. Active contraction of muscles can happen with fixed joints at any angle (isometric) or free rotating joints either with the contraction (concentric) or against the contraction (eccentric). This defines a complex state-function space for muscle in which there are numerous muscle lengths with the same joint angle but different activations, and numerous muscle lengths which have the same activation but different joint angles. If joint angles and neural drive are two independent inputs to muscle which contribute to state, some others are pressure from adjacent muscles, pain, inflammation, temperature, fatigue and historical states (state transitions).

The current state of the art technology does not provide a solution to measuring specific muscle states non-invasively. Surface electromyography (EMG) can non-invasively measure active contraction in superficial muscles, but measurements are noisy (need filtering), subjective, and correlations with measured muscle force are entirely dependent on the position of the electrode. Intramuscular EMG can provide invasive (needles or fine wires inserted through skin and muscle) measurements of active contractions in deep muscles. EMG cannot measure passive tension in the muscle and has many other well-known problems [1]. Dynamometry can provide non-invasive gross measures of passive or active forces acting on a joint and therefore can provide gross estimates of the contribution of groups of muscles crossing that joint. Dynamometry is therefore not muscle specific and cannot resolve passive and active tensions within specific muscles. Supersonic Shear wave Imaging (SSI) can provide non-invasive estimates of the regional stiffness within cross-sectional areas of specific muscles, where such measures have been shown to correlate well with measured/estimated muscle force [2]. SSI cannot resolve active force (produced from within the muscle from active of motor unit firing) from passive force (resulting from joint rotation or pressure from adjacent muscles), and

†This paragraph of the first footnote will contain the date on which you submitted your paper for review. It will also contain support information, including sponsor and financial support acknowledgment. For example, "This work was supported in part by the U.S. Department of Commerce under Grant BS123456".

The next few paragraphs should contain the authors' current affiliations, including current address and e-mail. For example, F. A. Author is with the National Institute of Standards and Technology, Boulder, CO 80305 USA (e-mail: author@ boulder.nist.gov).

S. B. Author, Jr., was with Rice University, Houston, TX 77005 USA. He is now with the Department of Physics, Colorado State University, Fort Collins, CO 80523 USA (e-mail: author@lamar.colostate.edu).

T. C. Author is with the Electrical Engineering Department, University of Colorado, Boulder, CO 80309 USA, on leave from the National Research Institute for Metals, Tsukuba, Japan (e-mail: author@nrim.go.jp).

correlations with measured force are subjective requiring calibration to person-specific maximum voluntary contraction (MVC). SSI requires that the ultrasound scanning plan be in line with the muscle fibers, else the theory breaks down [3]. Finally, SSI has a low sampling rate of about one sample per second [2].

There have been many previous attempts to analyse the features of skeletal muscle via ultrasound and yet this long standing medical need is still unmet. Previous attempts have all made assumptions as to what the descriptive features of active and passive tension are when observing muscle state within ultrasound; muscle thickness/cross-sectional area [4], muscle fascicle orientation/curvature [5]–[7], and muscle length [8], [9]. These previous attempts are either too presumptuous, too low dimensional, and/or do not sample the state-function space comprehensively. In this paper, we propose a new alternative approach for measuring states (torque, active contraction, passive shortening/lengthening) of individual muscles directly from ultrasound images using machine learning. The methods we have developed are applicable to standard frame-rate (25Hz) b-mode ultrasound imaging for numerous reasons, not least it is ubiquitous in a clinical and research sense, non-invasive, cost-effective, has minimal exclusion criteria, and is portable. Ultrasound can very easily image deep structures including deep muscles within the body at very practical frame rates (25-100+*Hz*). The focus of this paper is on the human calf muscles since they are of interest, with a dense research track record and a variety of previously developed computational methods.

## II. RELATED WORK

Although ultrasound has many clear benefits it is hard to analyze the information content [10]. With respect to skeletal muscle, research has predominantly been focused on extraction of intuitive low-resolution features such as pennation angle, and muscle thickness/cross-sectional area [4], [11]–[14]. Hodges and others [4] highlighted the potential of ultrasound for analyzing specific muscles without cross-talk (an artifact of EMG) from adjacent muscles and they compare low dimensional features with EMG measurements from four specific muscles undergoing isometric contractions. Among many conclusions, their most relevant findings were that from initial conditions large changes in the low dimensional features are associated with small changes in EMG, and small changes in low dimensional features are associated with large changes in EMG. They conclude from this that ultrasound would be good for measuring small activations (4-20% maximum voluntary contraction) but not large activations. If we assume that the main finding is true (i.e. large activations are associated with small changes in state) this merely means that there is a non-linear relationship between muscle state and contractile force, which does not mean that ultrasound should be limited to small activations. We must primarily consider measurement noise arising from human error a limiting factor for discrimination at higher forces, and then we must consider the fundamental limitation of using low-dimensional predicated features.

Rana and others [5] attempted to address the problem of subjectivity when measuring muscle fascicle orientation and curvature by developing a computational approach. After filtering their images with a vessel enhancement filter, they applied two methods to ultrasound images of the vastus lateralis muscle. The first method was the Radon transform which gave the main orientation of the visible fascicles. The second method was to convolve the images with orientated Gabor wavelet filters, where the maximum convolution at each pixel reveals the orientation for that location. The mean angle obtained from the wavelet convolutions reveals the dominant orientation over all of the fascicles. They evaluate their methods on synthetic data with known orientations, reporting accuracy to 0.02°. They also acquired manually digitized fascicle orientations from 10 operators, which revealed very large subjectivity between operators such that comparisons with the computational methods were not possible. They do not use automatic segmentation and fascicle regions were manually selected. The main problem with this approach is that there is no built-in discrimination of what is considered to be a fascicle; fascicles are vessel-like structures which exhibit a bright to dark to bright pixel intensity pattern, and there are many objects within an image which exhibit this pattern such as blood vessels, connective tissues, and nerves. These other structures can and often do present at different orientations from the fascicle field which can cause local errors in the measurement. It is currently not known if fascicle orientation is sufficient to delineate active and passive muscle states, or extract forces either within a single person or generalized outside a population.

For some time, the dominant paradigm was to track the motion of local structures visible within the image plane with a view to perhaps predicting and tracking muscle length; this is known as speckle, or feature tracking. The main flaw with this approach is tracking drift; where the local tracking error accumulates over time due to noise and other effects and the absolute position is lost. Loram and others [8] applied a cross-correlation feature tracking technique to two specific muscles (superficial and deep) in the lower leg. Loram demonstrated for the first time that ultrasound can be used to measure completely different muscle lengths in two adjacent muscles during the same task. Of the many conclusions, the relevant findings were that cross-correlation tracking fails for arbitrarily large movements, and that the unregulated tracking of point features results in tracking drift. The latter confirms the findings of an earlier study by Yeung and colleagues [15], [16] in which the authors address the tracking drift problem and conclude that it is a consequence of features leaving or entering the image plane making them inherently impossible to directly track with pure feature tracking methods.

The techniques of Loram and others [8] were improved upon by the use of a more robust tracking method; the Kanade Lucas-Tomasi (KLT) feature tracking method [17], [18], [9]. Loram and others mentioned that tracking failed for arbitrary large motions, but the KLT solved that problem by use of pyramid levels which could track at coarse detail (top pyramid level) for large motions, and refine tracking at each subsequent pyramid level. Darby and others [9] not only improved tracking of local features, they automated the entire analysis by applying active shape models (ASM) [19] and using Eigen features [18] interpolated at automatically placed grid points within the muscle belly using Delaunay triangulation. They reported inconsistent segmentation results with automatic initialization, and quite accurate results ($0.3mm$) with manual initialization. With respect to tracking, although they show robustness for larger movements, they found more drift than the cross-correlation method for small movements, and they reported large feature dropout (large discrepancies between texture patches in sequential images causing termination of tracking of that feature). The feature dropout they experienced resulted from out of plane motion, where features leave or enter the plane suddenly and cannot possibly be tracked using these methods.

Naturally we can now discuss regulated tracking methods [16], [7]. Regulated tracking within this domain is closely related to feature engineering; some presupposition about the information content is made and then a technique is developed to automatically measure that information, which is then used to regulate spurious tracking points. There is then a hierarchical dependency of feature tracking on the measured parameters, and the measured parameters on the quality of the data. Further, there are not always intelligible features within the muscle which can be used for regulation. The neck for example contains six bilateral muscle layers and when imaged simultaneously fascicles and other internal structures are invisible, only the muscle boundary and a random-deterministic internal speckle pattern are present. That speckle pattern can be tracked [20], but it would be difficult to formulate a regulated version of the tracking without some complex model of the speckle structure or neck muscle shape and mechanics.

The recent development [21]–[29] of a class of methods known collectively as deep learning (DL) provide a framework for understanding the content of ultrasound images of muscle in relation to measured data (EMG, torque, angle). DL is a technique for building ANN representations of data in a layer-wise fashion, where each layer models increasingly abstract/complex features of the data, facilitating modeling of complex features without *a priori* assumptions of the descriptive features. ANNs can learn nonlinear functions to map data (images) to labels (EMG, torque, joint angle). Even without many or any labels (which may often be the case with respect to deep muscles) features can be extracted using generative models such as restricted Boltzmann machines [21], [30], deep belief networks [31], deep autoencoders/autoassociators [32]–[34], or more recently generative adversarial networks (GAN) [35]. Those features can then either be directly analyzed (using statistics or distance metrics), or re-mapped to relatively few labels. If large volumes of labeled data exist, a CNN can be trained directly on the data to predict the labels, which can be continuous or discrete. CNNs work very well for understanding the content of static images [36] or speech [37], and more recently very deep CNNs known as residual networks (ResNet) have surpassed human-level performance [38] on the same image recognition task as [36]. CNNs have also demonstrated the ability to track local motion [39], which means that unlike standard feature tracking, a CNN can measure the dynamic state of local features, while simultaneously having access to the static state (or pose). One could argue for the need to measure historical states with short and/or long-term dependencies perhaps with recurrent networks, but considering this is an initial investigation of new ground they are not a sensible option since they are comparatively difficult to train and would not easily generalize to different ultrasound acquisition rates (e.g. ultrafast ultrasound $> 1000Hz$ *vs* standard 25-100Hz).

In order to establish a firm but accessible benchmark we investigate the application of standard deep CNNs, rather than recurrent or very deep ResNets, with a view to extending the current work in future research. Our CNNs are compared to a variation of the Darby method [9] which is fundamentally feature tracking with a fully connected feed-forward neural network on top. In addition, we used established visualization techniques to attempt to understand and interpret the models which we generate [40]. We show that these methods can be used to visually understand mechanical and functional differences between active and passive skeletal muscle.

III. METHODS

*A. Data Acquisition*

Ultrasound data were recorded from 19 participants (6 female, ages: $30 \pm 7.7$) during dynamic standing tasks. Participants stood upright on a programmable/controllable foot pedal system during three tasks while strapped at the hip to a backboard. During the tasks, we recorded calf muscle (GM) activation using electromyography (EMG), ankle joint angle/torque, all at $1000Hz$, and ultrasound of the GM at $25Hz$. Three distinct tasks were designed to explore the state-function space of muscle:

*1) Isometric*
The pedal system was fixed at a neutral angle (flat feet), and participants observed an analog oscilloscope. On the oscilloscope, we displayed side by side a dot representing the ampli-tude of their filtered EMG, and a dot representing the amplitude of a fabricated signal (see section III. C.). Participants were asked to contract their calf muscles by pushing down their toes while simultaneously keeping their foot in full contact with the static pedals.

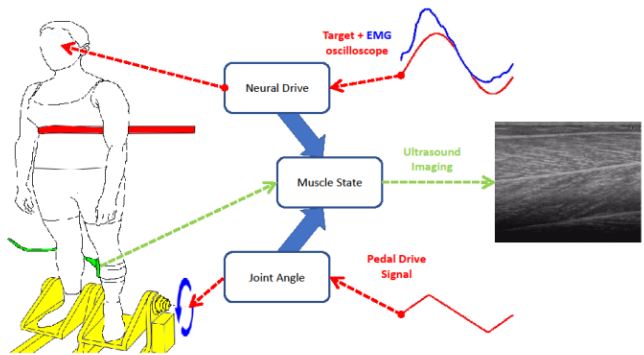

**Figure 1. Experimental setup.** In the figure, a participant stands upright on a foot pedal system (yellow highlight), while strapped (red strap) at the chest to a backboard, observing an oscilloscope screen opposite. The ultrasound probe (green highlight) is attached to the left lower leg next to a wireless EMG sensor, imaging the medial gastrocnemius (MG) and soleus muscles (grayscale image on the right). A target signal is shown on the screen (red line) along with EMG feedback (blue line) which the participant must match using contractions of the MG. A signal (red line) independently drives the foot pedal system about the rotational axis (blue circular arrows0, manipulating the ankle joint which the MG crosses.

*2) Passive*

Participants observed an analog oscilloscope. On the oscilloscope, we displayed side by side a dot representing the amplitude of their filtered EMG. Participants were asked to monitor and minimize any EMG activity be relaxing their muscles. The pedal system was driven using a fabricated signal (see section III. C.). Participants were asked to allow their ankle to rotate and keep their feet in full contact with the moving pedals.

*3) Combined*

The pedal system was fixed at a neutral angle (flat feet), and participants observed an analog oscilloscope. On the oscilloscope, we displayed side by side a dot representing the amplitude of their filtered EMG, and a dot representing the amplitude of a fabricated signal (see section III. C.). The pedal system was simultaneously driven using a fabricated signal (see section III. C.). Participants were asked to allow their ankle to rotate and keep their feet in full contact with the moving pedals.

All trials were 190 seconds in length which consisted of 10 seconds of neutral standing (i.e. no signals were used to move the pedals or the dot on the screen), followed by 180 seconds of trial. Data were collected in the ranges of 0.0481V, 100.182Nm, and 12.371° (*c.* 3.1° dorsiflexion, 9.3° plantarflexion) for EMG, torque, and joint angle respectively.

### B. Designing the Labels

Two signals were designed to manipulate active and passive muscle input factors, active contraction and passive joint rotation respectively. Both signals were derived from the following bases:

$$a = \sin\left(0.4t\pi - \frac{\pi}{2}\right),$$
$$b = \sin\left(0.5t\pi - \frac{\pi}{2}\right), \quad (1)$$
$$c = \sin\left(\sin\left(t\frac{\pi}{30} - \frac{\pi}{2}\right)30\pi - \frac{\pi}{2}\right).$$

The dot on the screen used to guide participants to contract their muscles was constructed using the following rules: 1) For the first 10 seconds signal *a* was used, and every 10 seconds thereafter we alternated between signals *a* and *b*. 2) After 30 seconds signal *c* was used, and every 30 seconds thereafter, either signal *a* or *b* was used depending on the first rule. The pedals were driven using the same bases with the following different rules: 1) For the first 20 seconds signal *a* was used, and every 20 seconds thereafter we alternated between signals *a* and *b*. 2) After 60 seconds signal *c* was used, and every 60 seconds thereafter, either signal *a* or *b* was used depending on the first rule. The signals were designed to produce transient correlations, de-correlations, and anti-correlations to maximize exploration of the state-function muscle space. The correlation

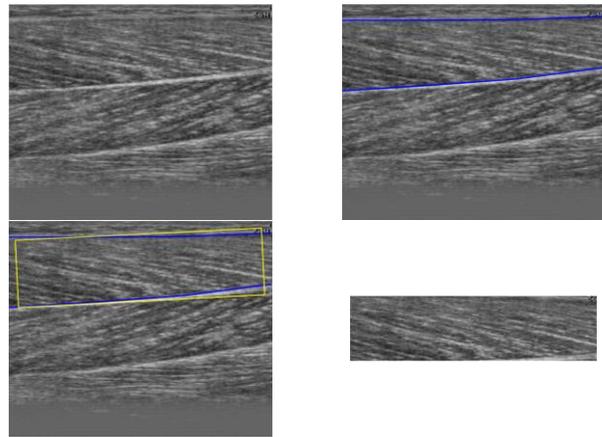

**Figure 2. Segmentation and region extraction pipeline.** The GM muscle in the raw ultrasound (tope left) is segmented (blue lines, top right) using the technique described in [41] (with direct manual annotations for training rather than MRI registration). After every image in the sequence is segmented, a region of interest is defined (yellow box, bottom left) using the mean segmentation over the sequence. Finally, the region is extracted in a standardized (rotating to 0° angle) image matrix.

of the two independent signals was $r = 0.33, p = 0$ (Pearson), and $r = 0.34, p = 0$ (Spearman).

Simulink (Matlab, R2013a, The MathWorks Inc., Natick, MA) was used to interface with the lab equipment (pedal system and EMG), and for video synchronization a hardware trigger was used to initiate recording at the start of each trial.

### C. Segmentation and Region Extraction

For segmentation and region extraction we used a fast and accurate muscle segmentation algorithm previously developed by our group [41]. That method enabled normalization of the gastrocnemius muscle such to reduce the computational dimensions and complexity while simultaneously maximizing the spatial resolution. The segmentation also provided an opportunity to standardize the input by allowing extraction of a region orthogonal to the main axis (mean over the video sequence) of the muscle (see figure).

First, an expert annotated the internal boundaries of the medial GM muscle in 500 randomly selected images of which 100 were randomly selected for testing. After interpolating the annotations to a standard 40 point vector, a principal component model was constructed from the remaining 400 images. The component model was then used to construct a texture-to-shape dictionary with only 4 components ($> 90\%$). That dictionary was then used to give an approximate segmentation for each image in the dataset. That initial segmentation was then used to initialize a heuristic search routine using an ASM [19] constructed from just 10 principal components ($> 99\%$). The search was conducted at full resolution $\pm 10$ pixels about each contour point. For more detail see [41].

The entire dataset ($> 300{,}000$ images) was segmented and then a region of interest ($x \times y = 496 \times 120 \; pixels \approx 55.42 \times 14.67 \; millimeters$) was extracted orthogonal to the main orientation of the GM muscle (linear least square fit to mean segmentation over the sequence).

### D. Feature Tracking

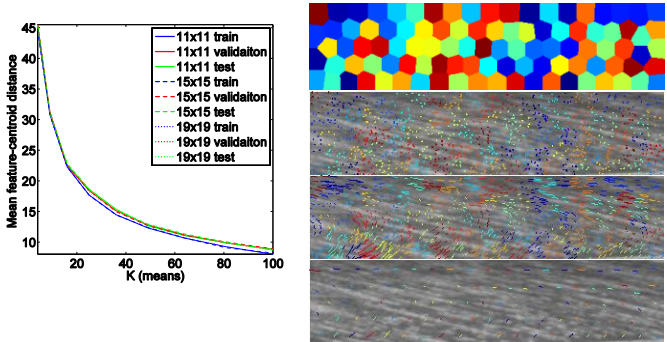

**Figure 3. Feature selection, k-means clustering, and tracking.** The left panel shows for increasing K (means) the mean centroid distance from K-means classified features for train, validation, and test sets, and 3 different KLT feature sizes (11, 15 and 19). The right panel shows the clustering and tracking process. Top: a tessellation diagram shows the 100 cluster regions. 2$^{nd}$ panel: 1000 features are selected using [18] and classified using the K-means clusters. 3$^{rd}$ panel: the motion of each feature into the next image is recorded. Bottom: the motions of features in the same class (same colour) are averaged and recorded as the representative motion of each cluster.

The first stage of feature tracking was selection of 'good' (corner) features [18]. Within the region of interest defined by the segmentation (see sub-section C), we selected the top 1000 Eigen features, where the number 1000 was chosen empirically (it was not always possible to select greater than approximately 1000 features using the Eigen features method [18]). Eigen features were selected in every image of every sequence. Following feature selection we took 500 equidistant (over trials and participants) images from the training set (see sub-section G) and used the K-means [42] algorithm to identify integration points from the selected features in those images. The integration points were used to average the motion of the features within a cluster and record a single motion for that region/cluster (see figure 2). We empirically chose 100 means which were used to classify every Eigen feature in every image of every sequence. Greater than 100 means caused empty clusters within the testing and/or validation sets, and less than 100 gave larger average centroid distances.

After K-means clustering the KLT [17] algorithm was used to track the motion of each of the Eigen features one image forward. The features belonging to the same class/cluster were averaged (mean) and recorded. The result was a matrix consisting of row vectors of 100 displacements ($x$, $y$), per image, per sequence (vector of 200 values). This was then used to train a variety of fully connected feed-forward neural networks (see sub-section F).

### E. Feed-forward Neural Network

After K-means clustering and integration of tracking points (previous section) we designed 3 feed-forward neural network architectures, with which to model the data and predict changes EMG, joint torque, and joint angle. The main design choices (or hyperparameters) in such networks are number of neurons in a layer, number of layers in a network, and they type of neuron transfer function. We treat the transfer function as a fixed hyperparameter and decided on the ReLU transfer function due to its popularity and success. We also treat the number of layers as a fixed hyperparameter and decided on 2 layers as a way of increasing complexity yet maintaining efficiency. The number of neurons per layer, however, was considered an important parameter and therefore we varied the number of neurons per layer in three models, A, B, and C, with 256, 512, and 1024 neurons, respectively.

These three models were trained with and without dropout ($p = 0.5$), with a learning rate of $1e - 5$, momentum of $9e - 1$.

### F. Convolutional Neural Network

When considering the choice of architecture our concern primarily was that the model was large enough to minimize the training error, and then the main concerns were in computation and generalization. Our strategy was to train a variety of models, exploring width (number of filters per convolutional layer) and depth (number of convolutional layers), with state of the art regularization (dropout), while evaluating performance on held out validation data. The architectures of the three modes were as follows:

| Model A | Model B | Model C | Model D |
|---------|---------|---------|---------|
| c-16    | c-36    | c-49    | c-64    |
| **p-2×2** | **p-2×2** | **p-2×2** | **p-2×2** |
| c-36    | c-64    | c-81    | c-121   |
| **p-2×2** | **p-2×2** | **p-2×2** | **p-2×2** |
| c-64    | c-121   | c-144   | c-169   |
| **p-2×2** | **p-2×2** | **p-2×2** | **p-2×2** |
| c-121   | c-169   | c-324   | c-225   |
| **p-2×2** | **p-2×2** | **p-2×2** | **p-2×2** |
| c-169   | c-225   | c-484   | c-289   |
| **p-2×2** | **p-2×2** | **p-2×2** | **p-2×2** |
| c-225   | c-289   | fc-1024 | c-361   |
| **p-2×2** | **p-2×2** | fc-1024 | **p-2×2** |
| c-289   | c-361   | fc-3    | c-441   |
| **p-2×2** | **p-2×2** |         | **p-2×2** |
| fc-1024 | fc-1024 |         | fc-1024 |
| fc-3    | fc-3    |         | fc-3    |

where the prefix c denotes convolutional layer, the prefix p denotes max pooling layer, the prefix fc denotes fully connected layer, and the number shown is the number of filters/neurons in the layer. The weight matrices associated with each convolutional filter were 2×2 in every layer except for the input layer which was connected to two sequential images in the form $n\times n\times 2$, where we varied $n$ during cross validation (sub-section G).

### G. Training and Cross Validation

To train our models we minimized the mean square error (MSE) between the model and the labels (change in EMG, joint torque, and joint angle) using stochastic online gradient descent (i.e. batch size of one). All images (or KLT features) and labels were normalized having zero mean and unit variance. A learning rate of $1e - 5$ was empirically chosen for KLT and CNN models, with momentum of $9.5e - 1$. ReLU units were used in all layers except the output layer which was linear. Prior to training all biases were initialized to 0, and all weights were initialized using a variation of the Xavier initialization [43],

$var(w) = \frac{1}{fan\_in}$.

The validation error was measured during training periodically to allow selection of optimal models. Cross validation was used with a test set of one held out participant (12500 samples) and a validation set of one held out participant (12500 samples). The validation set was used to choose the optimal model and the testing set was used to evaluate generalization performance. The same participants were used in cross validation in the CNN models and the KLT neural network models. The testing and validation sets were not used to train any of the models. To regularize our models we used dropout ($p = 0.5$) in the top 1, 2 or 3 fully connected layers (in both CNN and fully connected networks). Early stopping was used where the model with the lowest validation error was taken after the validation error did not decrease for more than 5 error evaluations.

**Table 1. Cross validation results.** This table shows the results for the model variations of the 2 different methods (KLT/CNN) in the form of mean square error (MSE). The first column shows the model being tested (asterisk indicates optimal validation), where model just means architecture (connectivity, number, type of layers, and number of filters/neurons) in each layer (i.e. complexity). The filter/window size column shows respectively the convolution filter size on the input layer, and the KLT feature/window size for a given model. For convolutional layers greater than 7 we doubled the number of layers between pooling layers (e.g. for 12 layers, we used 6 convolutional layers with an extra layer of the same number of filters between max pools) The dropout column shows how much dropout was used in terms of the number of layers from the output to which it was applied – the dropout coefficient was always 0.5.

| method | model | pool. layers | conv. layers | learning rate | filter/window size (input) | dropout layers | train MSE | validation MSE | test MSE |
|---|---|---|---|---|---|---|---|---|---|
| Raw Images + CNN | A | 6 | 6 | 1e-5 | 3x3 | 0 | 0.303 | 0.281 | 0.480 |
| | A | 6 | 6 | 1e-5 | 3x3 | 1 | 0.307 | 0.291 | 0.427 |
| | A | 6 | 6 | 1e-5 | 3x3 | 2 | 0.327 | 0.269 | 0.381 |
| | A | 5 | 10 | 1e-5 | 3x3 | 2 | 0.343 | 0.312 | 0.395 |
| | A | 6 | 12 | 1e-5 | 3x3 | 0 | 0.323 | 0.336 | 0.412 |
| | A | 6 | 12 | 1e-5 | 3x3 | 2 | 0.327 | 0.269 | 0.381 |
| | B | 6 | 12 | 1e-5 | 3x3 | 2 | 0.307 | 0.291 | 0.405 |
| | B | 7 | 14 | 1e-5 | 3x3 | 2 | 0.307 | 0.335 | 0.411 |
| | B | 7 | 7 | 1e-5 | 5x5 | 1 | 0.290 | 0.299 | 0.451 |
| | *B | 6 | 6 | 1e-5 | 5x5 | 2 | 0.316 | 0.253 | 0.403 |
| | B | 7 | 7 | 1e-5 | 5x5 | 2 | 0.321 | 0.288 | 0.423 |
| | D | 7 | 7 | 1e-5 | 5x5 | 2 | 0.350 | 0.283 | 0.397 |
| | B | 6 | 12 | 1e-5 | 5x5 | 2 | 0.329 | 0.312 | 0.416 |
| | C | 5 | 5 | 1e-3 | 7x7 | 2 | 0.287 | 0.340 | 0.522 |
| | C | 3 | 3 | 1e-5 | 11x11 | 2 | 0.287 | 0.276 | 0.453 |
| | B | 4 | 4 | 1e-5 | 11x11 | 2 | 0.373 | 0.309 | 0.457 |
| | C | 5 | 5 | 1e-3 | 11x11 | 2 | 0.303 | 0.323 | 0.499 |
| | B | 6 | 6 | 1e-5 | 11x11 | 2 | 0.373 | 0.261 | 0.438 |
| | C | 3 | 3 | 1e-5 | 11x11 | 3 | 0.341 | 0.318 | 0.457 |
| | C | 5 | 5 | 1e-5 | 11x11 | 3 | 0.285 | 0.281 | 0.457 |
| | B | 6 | 6 | 1e-5 | 11x11 | 3 | 0.440 | 0.280 | 0.453 |
| | D | 7 | 7 | 1e-5 | 11x11 | 3 | 0.322 | 0.297 | 0.451 |
| KLT + ANN | A | - | - | 1e-5 | 11x11 | 0 | 0.360 | 0.289 | 0.425 |
| | B | - | - | 1e-5 | 11x11 | 0 | 0.331 | 0.307 | 0.414 |
| | C | - | - | 1e-5 | 11x11 | 0 | 0.339 | 0.304 | 0.433 |
| | *A | - | - | 1e-5 | 15x15 | 0 | 0.323 | 0.286 | 0.412 |
| | B | - | - | 1e-5 | 15x15 | 0 | 0.379 | 0.293 | 0.442 |
| | C | - | - | 1e-5 | 15x15 | 0 | 0.348 | 0.295 | 0.416 |
| | A | - | - | 1e-5 | 19x19 | 0 | 0.341 | 0.302 | 0.403 |
| | B | - | - | 1e-5 | 19x19 | 0 | 0.340 | 0.295 | 0.424 |
| | C | - | - | 1e-5 | 19x19 | 0 | 0.311 | 0.320 | 0.400 |
| | A | - | - | 1e-5 | 11x11 | 1 | 0.405 | 0.313 | 0.465 |
| | B | - | - | 1e-5 | 11x11 | 1 | 0.431 | 0.327 | 0.499 |
| | C | - | - | 1e-5 | 11x11 | 1 | 0.420 | 0.310 | 0.491 |
| | A | - | - | 1e-5 | 15x15 | 1 | 0.403 | 0.299 | 0.495 |
| | B | - | - | 1e-5 | 15x15 | 1 | 0.388 | 0.290 | 0.469 |
| | C | - | - | 1e-5 | 15x15 | 1 | 0.416 | 0.292 | 0.497 |
| | A | - | - | 1e-5 | 19x19 | 1 | 0.408 | 0.301 | 0.472 |
| | B | - | - | 1e-5 | 19x19 | 1 | 0.397 | 0.294 | 0.471 |
| | C | - | - | 1e-5 | 19x19 | 1 | 0.397 | 0.312 | 0.473 |
| | A | - | - | 1e-5 | 11x11 | 2 | 0.545 | 0.422 | 0.569 |
| | B | - | - | 1e-5 | 11x11 | 2 | 0.528 | 0.409 | 0.543 |
| | C | - | - | 1e-5 | 11x11 | 2 | 0.523 | 0.403 | 0.551 |

*H. Visualization of CNN model*

We use established methods [40] to construct visualizations of the hierarchical knowledge learned by the best CNN model. More importantly we use these methods to understand the mechanical properties of active and passive muscle changes. For each convolutional layer, we generate images which maximize the response (output) of individual neurons (filters), and also images of each neuron maximized at every spatial location of the input space (2 sequential ultrasound images. Finally, we generate images which maximize the response of active (EMG) and passive (joint angle) neurons. Images are produced by initializing the input to the trained CNN with zero mean and unit variance Gaussian noise. Then we compute a full forward pass. Then we create an error vector which is equal to the output of the layer of interest, plus a constant (1) at the unit we want to maximize. Then from the layer of interest we back-propagate that error through to the input layer, and we use the gradient to update the pixels (at a rate of (0.5). During updates we apply L2 regularization (0.05) as per [40]. This process is repeated with the new pixels for 100 iterations.

## IV. RESULTS

**Table 2. EMG test case results.**

| function | method | EMG MSE | NRMSE | RMSE | $R^2$ |
|---|---|---|---|---|---|
| combined | CNN | 1.379 | 0.319 | 0.0009V | 0.537 |
| | KLT + ANN | 1.463 | 0.299 | 0.0009V | 0.508 |
| isometric | CNN | 0.775 | 0.308 | 0.0007V | 0.522 |
| | KLT + ANN | 0.778 | 0.307 | 0.0007V | 0.520 |
| passive | CNN | 0.014 | - | 0.0001V | - |
| | KLT + ANN | 0.014 | - | 0.0001V | - |

**Table 3. Torque test case results.**

| function | Method | Torque MSE | NRMSE | RMSE | $R^2$ |
|---|---|---|---|---|---|
| Combined | CNN | 0.709 | 0.302 | 0.922Nm | 0.513 |
| | KLT + ANN | 0.679 | 0.317 | 0.902Nm | 0.533 |
| Isometric | CNN | 0.086 | 0.377 | 0.321Nm | 0.612 |
| | KLT + ANN | 0.089 | 0.365 | 0.328Nm | 0.597 |
| Passive | CNN | 0.055 | 0.277 | 0.256Nm | 0.477 |
| | KLT + ANN | 0.054 | 0.278 | 0.256Nm | 0.480 |

**Table 4. Joint angle test case results.**

| function | method | Joint Angle MSE | NRMSE | RMSE | $R^2$ |
|---|---|---|---|---|---|
| combined | CNN | 0.463 | 0.457 | 0.139° | 0.705 |
| | KLT + ANN | 0.443 | 0.469 | 0.136° | 0.718 |
| isometric | CNN | 0.011 | - | 0.022° | - |
| | KLT + ANN | 0.022 | - | 0.030° | - |
| passive | CNN | 0.131 | 0.709 | 0.074° | 0.915 |
| | KLT + ANN | 0.138 | 0.702 | 0.076° | 0.911 |

### A. Region Extraction

The segmentation technique we had previously developed was evaluated by manual annotation of 100 test images. Our concern here was not generalization but accuracy within the dataset. Our results showed that the segmentation was accurate to $0.16mm$ (> 99.9%) and segmented approximately 10 images per second.

### B. Modeling Muscle Function from State

The KLT + ANN method was able to predict changes EMG, torque, and joint angle to within 0.0007V, 0.57Nm and 0.09°, respectively. The CNN method was able to predict changes in EMG, torque, and joint angle to within 0.0006V, 0.58Nm and 0.09°, respectively. Comparison of the two different computational techniques revealed few qualitative/quantitative differences in performance. The KLT method was less discriminative of active and passive function in the isometric and passive cases (see figure, and table) for the test participant, yet showed a slight improvement over the CNN in the combined function case for torque and ankle angle predictions. The CNN method performed better than the KLT method when predicting EMG, and worse when predicting torque, while there was very little difference for ankle angle (see tables 1-3).

Analysis of CNN cross-validation results revealed that the most important factors for generalization were filter size in the input layer, learning rate, and number of convolutional and pooling layers; the width (number of filters per convolutional layer) of the network was less important. Network depth (number of conv./pooling layers) broadly improved performance, although adding too many pooling layers (i.e. so input dimensions to fc. layers reached $2\times1\times n$ where $n$ is the number of filters in the last layer) proved detrimental to generalization. The sizes of the filters in each convolutional layer remained fixed parameters at $3\times3$, except for the input layer, where we found smaller filter sizes improved generalization. The learning rate proved to be the factor with the greatest effect on generalization. Observations of the convolution ReLU response histograms during training revealed large dropout (dying ReLU) in many layers where the learning rate was large (1e-3), and was much more stable for lower learning rates (1e-5). Online training prevented use of adaptive learning rate algorithms. Finally, initialization of the weights was an important factor. Empirical experiments using initial weights sampled from a Gaussian distribution with unit variance and zero mean proved very difficult to train using ReLUs, but not such an issue for other unit types; the *tanh* unit was investigated but convergence was more of an issue than initial training. Using Xavier initialization our CNNs trained much more easily in just a few hours. All networks converged (diverged from the validation set) within 2.5 million weight updates.

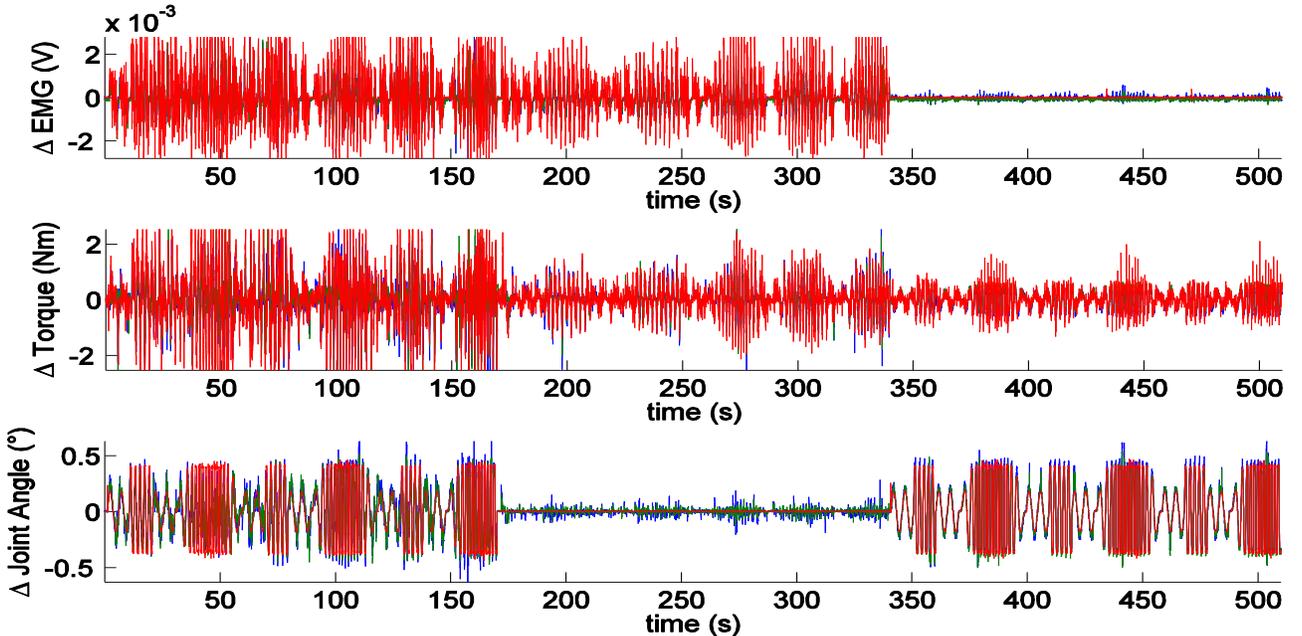

**Figure 4. Overview of time-series of test results.** This figure shows time-series mode predictions from KLT (blue) and CNN (green) models compared with the original data (red) for the held-out test participant. All data presented has had no filtering or temporal smoothing (except for raw EMG filtering to create labels prior to modelling); these are the raw model outputs. In general, the figure shows good tracking of all 3 signals, with the discriminative power becoming particularly apparent in the top and bottom panels where in the passive (time > 340) and isometric (time > 170, time < 340) cases both KLT and CNN accurately suppress responses to stimulus, with perhaps the CNN demonstrating the better performance.

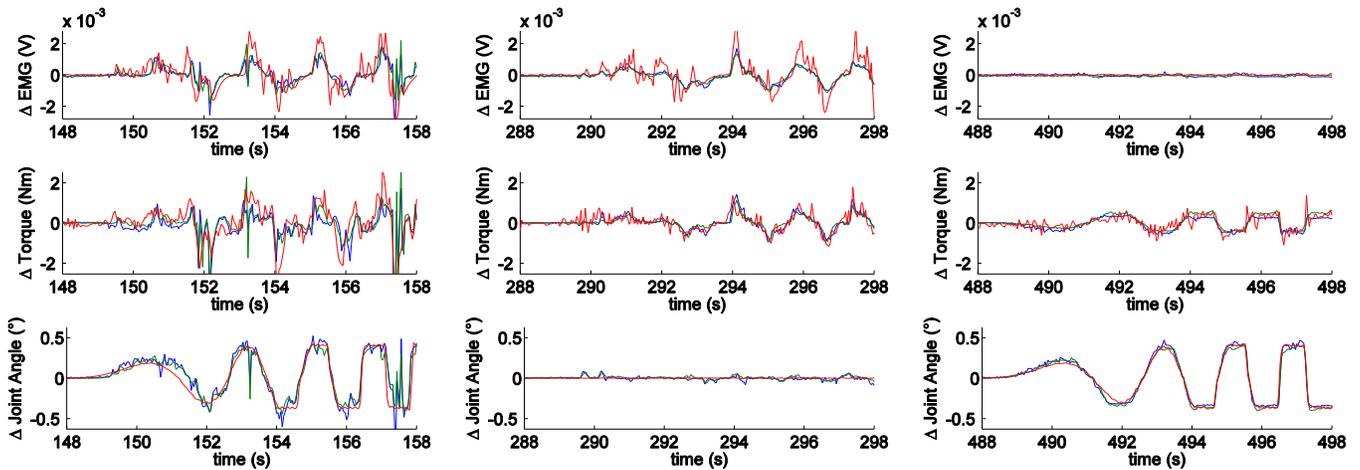

**Figure 5. Zoom of time-series test results.** This figure clearly illustrates the discriminative power of ultrasound for analysis of skeletal muscle. All data presented has had no filtering or temporal smoothing (except for raw EMG filtering to create labels prior to modelling); these are the raw model outputs. The KLT (blue) and CNN (green) give very similar responses, which in some respects validate one another; since the CNN uses the raw texture (i.e. no predefined features) for such a complex analysis, one would expect a bad model not to corroborate with one with previously well-defined features (KLT). The panel on the bottom right shows joint angle data being tracked and predicted very well by both CNN and KLT in the passive case, while isometric and combined cases show generally slightly weaker and less stable performance. Torque data is generally predicted very well considering this is a generalized (test) participant; again the combined case is weaker. EMG data as a continuous variable is predicted most unreliably, however the ability to discriminate active and passive motions is captured well.

Analysis of KLT results revealed very little difference between models. Reducing KLT feature size from 19×19 to 11×11 had little noticeable or measurable effect. The main factor with respect to generalization was model complexity. Smaller models generalized better. Dropout regularization had a negative effect on generalization, where 2 layers of dropout caused convergence at very high error.

## V. DISCUSSION

This manuscript details the first published result of its kind within its domain; i.e. generalized prediction of changes in muscle-specific torque, connecting joint angle, motor unit activity (EMG) directly from standard frame-rate (25Hz) b-mode ultrasound in combined functional conditions. This manuscript also demonstrates the successful application of CNNs to medical ultrasound outside the domain of classification. Our CNNs were trained with relative ease after empirical tests revealed good learning parameters. There is currently no benchmark with which to compare to, hence we implemented an existing state of the art technique based on KLT feature tracking, and with some parameterization standard feed-forward ANNs provided a relevant comparison. We have demonstrated state of the art performance with our CNNs with only marginally smaller errors than the KLT with ANN. All the literature on deep learning suggests that this benchmark can be improved upon with additional data. While our dataset of over 300,000 images may seem large, there were only 19 participants, and because 2 were held out, that leaves only 17 different muscle architectures, probe positions/orientations, and tissue compositions with which to generalize from. The fact that we have produced generalized predictions of torque from specific muscles in combined functional conditions from only 17 participants is testament to the feasibility of the methods to the problem of measuring torque from specific muscles non-invasively.

After model selection we applied a standard visualization technique [40] to gain insight into what our CNN had modeled with respect to active and passive muscle function. The technique works the same way as learning in a CNN, where the input is initialized (either with a pair of images or random noise – we used noise) and instead of learning (in the active case) the weights which predict an EMG burst, the error gradient is used to learn the images which predict an EMG burst. The result was 2 image pairs representing a single motion; one motion for active and one motion for passive. The active reconstruction depicted a shearing motion (the superficial and deep parts of the muscle moved right and left respectively), while the passive motion illustrated a broadly uniform left linear translation. The ability to produce these graphics from CNNs is a particularly powerful paradigm, especially in the domain of medical image analysis. For example, if one were to construct a model of muscle function from a complex system like the posterior neck, which consists of 6 bilateral muscle layers, without segmentation, in theory this technique could provide localization of abnormal contractions of the kind that happen to people with cervical dystonia.

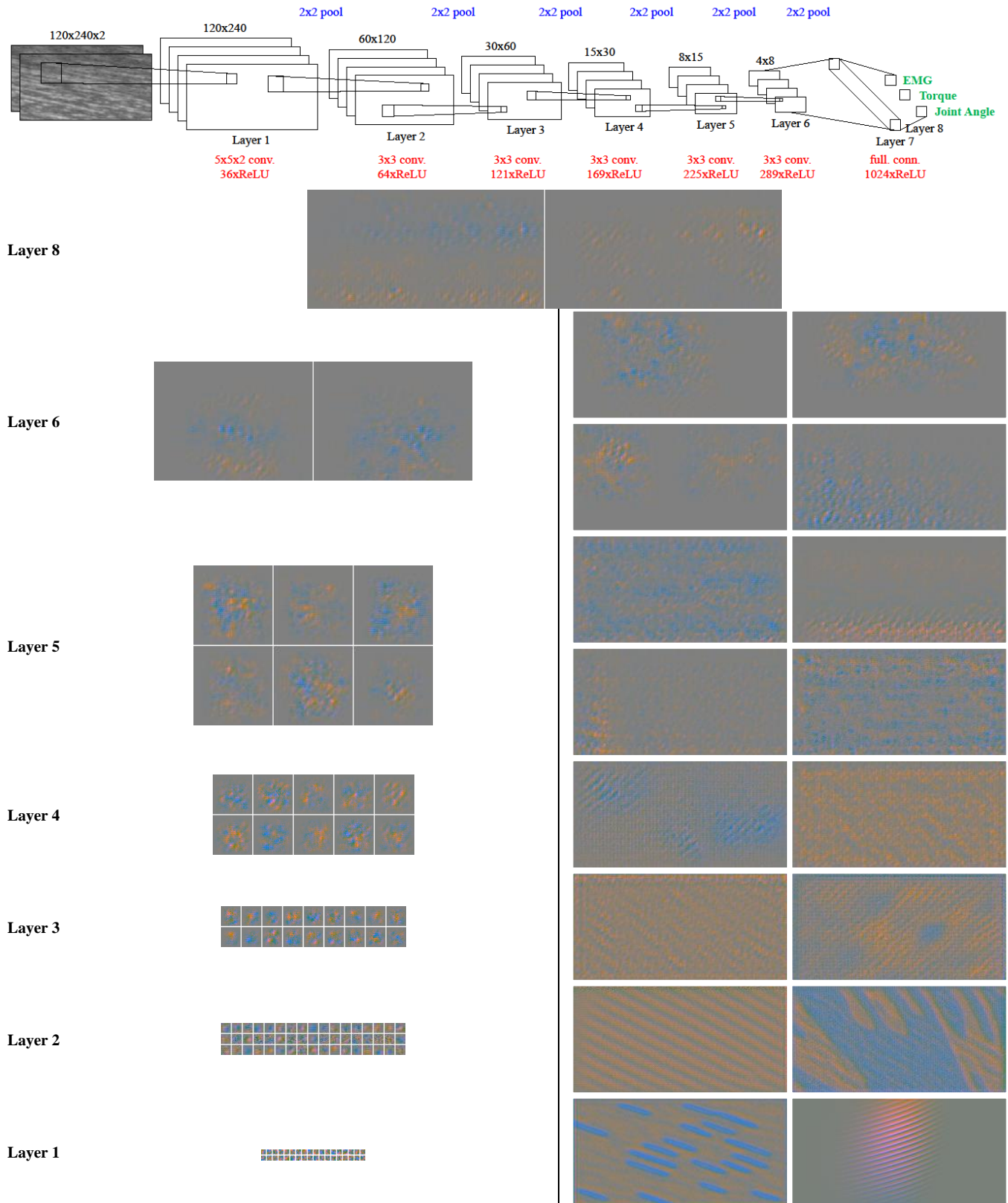

**Figure 5. CNN architecture and visualization.** This figure illustrates the knowledge learned by the best CNN model for each convolutional layer, and the final output/prediction layer (layer 8). The top panel shows the network schematic. In the other panels, each image represents the spatial texture information from two consecutive images (input), therefore to aid visualization each of the two images have been encoded as a separate color channel, where blue is the image at time $t$ and green is the image at time $t$-1. The panel on the left shows, for each layer, the optimized image of a random subset of the receptive fields of individual neurons (i.e. maximizing a neuron over one spatial pixel) at actual scale in pixels. The panel on the right shows for each convolutional layer the optimized image of selected neuron maps (i.e. maximizing a neuron over every spatial pixel) at actual scale in pixels. The left panel illustrates a diverse range of motion filters which look like a optical flow fields, except here the knowledge presumably encodes complex trans-planar texture changes as well as localized affine transformations. The right panel shows some of the abstract patterns that convolutional layers 'like to see' which become more localized/specialized in the superficial layers. The layer 8 graphics show the images which respectively maximize the EMG (active) and joint angle (passive) neurons, where the active map shows opposing directions of motion between the superficial and deep parts of the muscle, and the passive map shows homogeneous horizontal motion.

The success of both methods opens up a new domain for research and development, namely low-cost personalized non-invasive measurement of torque from deep skeletal muscles. This has clinical relevance to many musculoskeletal diseases like motor neurone disease for measurement and monitoring of twitches and dystonia for measurement, targeting and treatment of abnormal muscle contractions. The CNN method however opens up an additional domain over methods like the KLT, namely drift-less tracking of functional muscle states for drift-less prediction of muscle specific torque. Currently both methods presented here track changes in muscle state and when accumulated over time they drift from the absolute measurement. We have not addressed this issue here, since we address a different problem. Feature tracking methods (KLT) drift for a variety of fundamental reasons (noise, viewpoint variation, occlusion/ object orientation variation, etc..), and a KLT tracker has no model of the underlying texture. Our CNNs observed patches of texture in 2 adjacent images in a video sequence and learned to map changes in texture to changes in muscle functional states (EMG, torque, and angle). While the CNNs observed changes in texture as per the nature of the problem, they also had access to the absolute state in both images and could therefore potentially model absolute states and map them to absolute changes in muscle functional states (i.e. the difference between a muscle at rest and a muscle in some arbitrary state of torque output) – this is not possible with the KLT. Our results suggest this ought to be possible within some arbitrary range since we demonstrated successful tracking within the experimental range. We could feasibly predict that a KLT approach may work up until some maximum velocity due to the unconstrained nature of the algorithm and the motion of features outside the ultrasound image plane which are fundamentally not trackable, while the CNN could work beyond this range and perhaps for any conceivable range of motion because of the intrinsic ability to build models of range of motion, rather than tracking nearest matching texture patches.

The consistency of both KLT and CNN over all models and parameters was likely due to the near-perfect normalization of the data using our existing segmentation technique [41]. Accurate segmentation of skeletal muscle for modeling purposes is extremely uncommon. The benefits of segmentation to our approaches are that over-fitting is more difficult (i.e. generalization is encouraged), and also that there can be no doubt that the measurements we extract are from specific muscles. We propose that it is possible to recreate this analysis without segmentation; KLT tracking would provide some degree of functional segmentation for an ANN system, as would some texture differences for the CNN to some degree. Also, CNNs inherently cope well with translation of features through the max-pooling layers and weight sharing during convolution. However, we recommend segmentation where possible to validate measurements from specific muscles.

VI. CONCLUSIONS

In this paper we have presented a novel experiment for the generation of thousands of accurately labeled muscle ultrasound images for modeling functional muscle states using ultrasound. We have presented the first generalized prediction of specific muscle EMG, torque, and joint angle from standard frame-rate b-mode ultrasound for combined functional cases. Existing methods rely on simplistic measures in isolated cases (isometric only, or passive only) which do not generalize and have negligible practical application. We have demonstrated the efficacy of CNNs to this domain, which opens up a whole new line of research, namely deep learning applied to skeletal muscle ultrasound. The work presented here could realistically have practical applications in sport and performance biomechanics, and clinical applications in rehabilitation, diagnosis and monitoring of cervical dystonia and motor neurone disease. Although we have not demonstrated application to deep muscles, the techniques presented here are easily transferable to deep muscles. Future research will focus on increasing the population and functional range (larger torques and joint rotations with additional joint variables – i.e. the knee) in our dataset. We will also focus on extending the current research to absolute measurement of torque in multiple muscles both with and without segmentation.

VII. REFERENCES

[1] K. Türker, "Electromyography: some methodological problems and issues," *Phys. Ther.*, vol. 73, no. 10, pp. 698–710, 1993.

[2] F. Hug, K. Tucker, J. L. Gennisson, M. Tanter, and A. Nordez, "Elastography for Muscle Biomechanics: Toward the Estimation of Individual Muscle Force," *Exerc. Sport Sci. Rev.*, vol. 43, no. 3, pp. 125–133, 2015.

[3] S. F. Eby, P. Song, S. Chen, Q. Chen, J. F. Greenleaf, and K. N. An, "Validation of shear wave elastography in skeletal muscle," *J. Biomech.*, vol. 46, no. 14, pp. 2381–2387, 2013.

[4] P. W. Hodges, L. H. M. Pengel, R. D. Herbert, and S. C. Gandevia, "Measurement of muscle contraction with ultrasound imaging," *Muscle and Nerve*, vol. 27, no. 6, pp. 682–692, 2003.

[5] M. Rana, G. Hamarneh, and J. M. Wakeling, "Automated tracking of muscle fascicle orientation in B-mode ultrasound images," *J. Biomech.*, vol. 42, no. 13, pp. 2068–2073, 2009.

[6] A. I. L. Namburete, M. Rana, and J. M. Wakeling, "Computational methods for quantifying in vivo muscle fascicle curvature from ultrasound images," *J. Biomech.*, vol. 44, no. 14, pp. 2538–2543, 2011.

[7] J. Darby, B. Li, N. Costen, I. Loram, and E. Hodson-Tole, "Estimating skeletal muscle fascicle curvature from B-mode ultrasound image sequences," *IEEE Trans. Biomed. Eng.*, vol. 60, no. 7, pp. 1935–1945, 2013.

[8] I. Loram, C. Maganaris, and M. Lakie, "Use of ultrasound to make noninvasive in vivo measurement of continuous changes in human muscle contractile length," *J. Appl. Physiol.*, pp. 1311–1323, 2006.

[9] J. Darby, E. F. Hodson-Tole, N. Costen, and I. D. Loram, "Automated regional analysis of B-mode ultrasound images of skeletal muscle movement.," *J. Appl. Physiol.*, vol. 112, no. 2, pp. 313–327, 2012.

[10] J. a. Noble and D. Boukerroui, "Ultrasound image

segmentation: a survey," *IEEE Trans. Med. Imaging*, vol. 25, no. 8, pp. 987–1010, 2006.

[11] C. N. Maganaris and V. Baltzopoulos, "Predictability of in vivo changes in pennation angle of human tibialis anterior muscle from rest to maximum isometric dorsiflexion," *Eur. J. Appl. Physiol. Occup. Physiol.*, vol. 79, no. 3, pp. 294–297, 1999.

[12] S. Ikegawa, K. Funato, N. Tsunoda, H. Kanehisa, T. Fukunaga, and Y. Kawakami, "Muscle force per cross-sectional area is inversely related with pennation angle in strength trained athletes.," *J. Strength Cond. Res.*, vol. 22, no. 1, pp. 128–31, 2008.

[13] O. M. Rutherford and D. A. Jones, "Measurement of fibre pennation using ultrasound in the human quadriceps in vivo," *Eur. J. Appl. Physiol. Occup. Physiol.*, vol. 65, no. 5, pp. 433–437, 1992.

[14] K. Häkkinen and K. L. Keskinen, "Muscle cross-sectional area and voluntary force production characteristics in elite strength- and endurance-trained athletes and sprinters," *Eur. J. Appl. Physiol. Occup. Physiol.*, vol. 59, no. 3, pp. 215–220, 1989.

[15] F. Yeung, S. F. Levinson, and K. J. Parker, "Multilevel and motion model-based ultrasonic speckle tracking algorithms," *Ultrasound Med. Biol.*, vol. 24, no. 3, pp. 427–441, 1998.

[16] F. Yeung, S. F. Levinson, D. Fu, and K. J. Parker, "Feature-Adaptive Motion Tracking of Ultrasound Image Sequences Using A Deformable Mesh," vol. 17, no. 6, pp. 945–956, 1998.

[17] C. Tomasi, "Detection and Tracking of Point Features," *Sch. Comput. Sci. Carnegie Mellon Univ.*, vol. 91, no. April, pp. 1–22, 1991.

[18] J. Shi and C. Tomasi, "Good Features to Track," in *Computer Vision and Pattern Recognition, 1994. Proceedings CVPR '94., 1994 IEEE Computer Society Conference on*, 1994, pp. 593–600.

[19] T. F. Cootes, C. J. Taylor, D. H. Cooper, and J. Graham, "Active Shape Models-Their Training and Application," *Computer Vision and Image Understanding*, vol. 61, no. 1. pp. 38–59, 1995.

[20] I. Loram, B. Bate, P. Harding, R. Cunningham, and A. Loram, "Proactive selective inhibition targeted at the neck muscles: this proximal constraint facilitates learning and regulates global control," *IEEE Trans. Neural Syst. Rehabil. Eng.*, vol. 4320, no. c, pp. 1–1, 2017.

[21] G. Mayraz and G. E. Hinton, "Recognaizing Handwritten Digital Using Hierarchical Products of Experts," vol. 24, 2, no. 2, pp. 189–197, 2002.

[22] Y. LeCun, Y. Bengio, G. Hinton, L. Y., B. Y., and H. G., "Deep learning," *Nature*, vol. 521, no. 7553, pp. 436–444, 2015.

[23] L. D. Le Cun Jackel, B. Boser, J. S. Denker, D. Henderson, R. E. Howard, W. Hubbard, B. Le Cun, J. Denker, and D. Henderson, "Handwritten Digit Recognition with a Back-Propagation Network," *Adv. Neural Inf. Process. Syst.*, pp. 396–404, 1990.

[24] A. Krizhevsky, I. Sutskever, and G. E. Hinton, "ImageNet Classification with Deep Convolutional Neural Networks," *Adv. Neural Inf. Process. Syst. 25*, pp. 1–9, 2012.

[25] K. He, X. Zhang, S. Ren, and J. Sun, "Deep Residual Learning for Image Recognition," *Arxiv.Org*, vol. 7, no. 3, pp. 171–180, 2015.

[26] V. Nair and G. E. Hinton, "Rectified Linear Units Improve Restricted Boltzmann Machines," *Proc. 27th Int. Conf. Mach. Learn.*, no. 3, pp. 807–814, 2010.

[27] G. Hinton, "Dropout : A Simple Way to Prevent Neural Networks from Overfitting," vol. 15, pp. 1929–1958, 2014.

[28] H. Noh, S. Hong, and B. Han, "Learning deconvolution network for semantic segmentation," in *Proceedings of the IEEE International Conference on Computer Vision*, 2016, vol. 11–18–Dece, pp. 1520–1528.

[29] G. E. Hinton, "Training products of experts by minimizing contrastive divergence," *Neural Comput.*, vol. 14, no. 8, pp. 1771–1800, 2002.

[30] P. Smolensky, "Information processing in dynamical systems: Foundations of harmony theory," *Parallel Distrib. Process. Explor. Microstruct. Cogn.*, vol. 1, no. 1, pp. 194–281, 1986.

[31] G. E. Hinton, P. Dayan, B. J. Frey, and R. M. Neal, "The 'wake-sleep' algorithm for unsupervised neural networks.," *Science (80-. ).*, vol. 268, no. 5214, pp. 1158–1161, 1995.

[32] D. C. Plaut and G. E. Hinton, "Learning sets of filters using back-propagation," *Comput. Speech Lang.*, vol. 2, no. 1, pp. 35–61, 1987.

[33] J. Masci, U. Meier, D. Cireşan, and J. Schmidhuber, "Stacked convolutional auto-encoders for hierarchical feature extraction," *Lect. Notes Comput. Sci. (including Subser. Lect. Notes Artif. Intell. Lect. Notes Bioinformatics)*, vol. 6791 LNCS, no. PART 1, pp. 52–59, 2011.

[34] G. E. Hinton and R. R. Salakhutdinov, "Reducing the Dimensionality of Data with Neural Networks\r," *Science (80-. ).*, vol. 313, no. 5786, pp. 504–507, 2006.

[35] I. Goodfellow, J. Pouget-Abadie, and M. Mirza, "Generative Adversarial Networks," *arXiv Prepr. arXiv ...*, pp. 1–9, 2014.

[36] a Krizhevsky, I. Sutskever, and G. Hinton, "Imagenet classification with deep convolutional neural networks," *Adv. Neural Inf. Process. Syst.*, pp. 1097–1105, 2012.

[37] G. Hinton *et al.*, "Deep Neural Networks for Acoustic Modeling in Speech Recognition," *IEEE Signal Process. Mag.*, no. November, pp. 82–97, 2012.

[38] K. He, X. Zhang, S. Ren, and J. Sun, "Delving Deep into Rectifiers: Surpassing Human-Level Performance on ImageNet Classification," *CoRR*, vol. abs/1502.0, 2015.

[39] P. Fischer, A. Dosovitskiy, E. Ilg, ..., and T. Brox, "FlowNet: Learning Optical Flow with Convolutional Networks," *Iccv*, p. 8, 2015.

[40] M. D. Zeiler and R. Fergus, "Visualizing and Understanding Convolutional Networks arXiv:1311.2901v3 [cs.CV] 28 Nov 2013," *Comput. Vision–ECCV 2014*, vol. 8689, pp. 818–833, 2014.

[41] R. Cunningham, P. Harding, and I. Loram, "Real-Time Ultrasound Segmentation, Analysis and Visualization of Deep Cervical Muscle Structure," *Trans. Med. Imaging*, vol. 62, no. c, 2015.

[42] J. MacQueen, "Some methods for classification and analysis of multivariate observations," *projecteuclidorg*, vol. 1, pp. 281–297, 1967.

[43] X. Glorot and Y. Bengio, "Understanding the difficulty of training deep feedforward neural networks," *Proc. 13th Int. Conf. Artif. Intell. Stat.*, vol. 9, pp. 249–256, 2010.